\documentclass[sensors,article,submit,pdftex,moreauthors]{Definitions/mdpi}

\usepackage{subcaption}

\firstpage{1}
\pubvolume{1}
\issuenum{1}
\articlenumber{0}
\pubyear{2026}
\copyrightyear{2026}
\datereceived{ }
\daterevised{ }
\dateaccepted{ }
\datepublished{ }

\Title{CAppleNet: A Coordinate Attention-Based YOLO11 for Real-Time Apple Stem Cavity and Calyx Detection}

\Author{Alican Güneş$^{1,2}$\orcidA{}, Robert Selim Hill$^{2}$\orcidB{}, Gökhan Hasanoğlu$^{2}$\orcidC{} and Murat Kurt $^{1,*}$\orcidD{}}

\AuthorNames{Alican Güneş, Robert Selim Hill, Gökhan Hasanoğlu and Murat Kurt}

\address{%
$^{1}$ \quad International Computer Institute, Ege University, \.{I}zmir, T{\"u}rkiye; alican.acg@gmail.com (A.G.); murat.kurt@ege.edu.tr (M.K.)\\
$^{2}$ \quad Tarend, \.{I}zmir, T{\"u}rkiye; alican.acg@gmail.com (A.G.); rs.hill@tarend.com (R.S.H.); g.hasanoglu@tarend.com (G.H.)}

\corres{Correspondence: murat.kurt@ege.edu.tr; Tel.: +90-232-311-3227 (M.K.)}

\abstract{Automated fruit grading, sorting, and orientation systems require real-time, accurate detection of fruit-level morphological landmarks rather than the fruit boundary alone. This paper presents a custom, fully self-collected and self-annotated dataset of 1506 apple images (1054 training, 301 validation, 151 test) captured under controlled, fixed-camera conditions and annotated for three target classes: fruit body, stem cavity, and calyx. A YOLO11s detector trained with default hyperparameters is adopted as the baseline, and a systematic, one-factor-at-a-time ablation study is conducted across input resolution, data augmentation policy, optimizer configuration, and lightweight architectural enhancement to identify the factors that most improve detection of these fine-grained landmarks under high inter-cultivar color and texture variability. Neither increased input resolution, nor any of six tested augmentation variants, nor manual optimizer tuning improved on the untuned baseline (F1 = 0.9047); only a same-slot architectural substitution of the backbone's C2PSA block with Coordinate Attention (CA) matched the baseline's overall F1 (0.9038) while reducing parameter count by 10.2\% and achieving the study's best mAP@0.5 (0.935) and best stem-cavity F1 (0.8217). A confidence-threshold sweep identified 0.15 as the deployment operating point for this model, and a confirmatory experiment combining the CA-swap architecture with the best-performing augmentation variant showed the two effects do not combine additively, reversing the architectural gain rather than compounding it. The resulting design recommendations --- the CA-swap architecture alone, at a 0.15 confidence threshold --- target real-time apple landmark detection for orientation-aware industrial sorting and alignment systems.}


\begin{document}

\section{Introduction}
\label{sec:intro}

Automated fruit grading, sorting, and orientation systems depend on the accurate, real-time localization of fruit-level morphological landmarks rather than the fruit boundary alone. For apples specifically, reliable detection of the stem cavity and calyx (flower end), together with the overall fruit body, is a prerequisite for orientation-aware packaging, defect inspection, and robotic handling, where misidentification of these regions directly degrades downstream mechanical and quality-control decisions.

Deep-learning-based object detectors have progressively replaced classical image-segmentation pipelines for fruit localization tasks \cite{Bargoti2017,Lin2019}. Early convolutional approaches demonstrated the feasibility of in-field and orchard-based fruit detection and yield estimation \cite{Bargoti2017,Tian2019,Kang2019}, while the introduction of the YOLO family provided a favorable accuracy--speed trade-off that made real-time deployment practical \cite{Bochkovskiy2020,Liu2020}. Building on this foundation, a substantial body of work has adapted successive YOLO variants specifically to apple detection: improved YOLOv3 and YOLOv5 variants for orchard and picking-robot scenarios \cite{Tian2019,Yan2021}, channel-pruned and lightweight YOLOv5/YOLOv4 architectures for fruitlet detection and complex backgrounds \cite{Wang2021,Zhang2022}, ShuffleNetV2-YOLOX and attention-augmented YOLOv5 (CA-YOLOv5) models for natural-environment robustness \cite{Ji2022,Yang2024}, and, most recently, lightweight embedded-system and improved YOLO11-based detectors such as YOLO-WAS \cite{OlguinRojas2025,Du2025}. Broader surveys of the YOLO family and its agricultural applications \cite{Alif2024,Lv2025} confirm a consistent trend toward progressively lighter, faster architectures without sacrificing detection accuracy.

Closest to the present work, Baek et al. proposed AS-YOLO, which combines a Ghost Bottleneck design with a global attention mechanism specifically for apple stem-region segmentation \cite{Baek2025}, confirming that fine-grained, sub-fruit landmarks such as the stem region require architectural consideration beyond whole-fruit bounding-box detection. However, this and related studies typically report a single, already-optimized architecture rather than a systematic, controlled comparison of the individual design choices that led to it. In particular, it remains unclear which factors --- data augmentation policy, input resolution, optimizer configuration, or lightweight architectural modules --- contribute most to detecting small, fine-grained landmarks such as the stem cavity and calyx, especially under an imaging setup in which camera distance and illumination are fixed, but fruit appearance (color, blemishes, bicoloration) varies substantially across cultivars.

To address this gap, the present study makes the following contributions:
\begin{itemize}
    \item \textbf{Custom, fully self-collected and self-annotated dataset:} Unlike studies relying on public, synthetic, or orchard-captured imagery \cite{Bargoti2017,Lin2019,Kang2019}, all images in this study were captured under controlled, fixed-camera conditions and manually annotated by the authors on the Roboflow platform, yielding 1506 images (1054 training, 301 validation, 151 test) across three classes: fruit body, stem cavity, and calyx.
    \item \textbf{A rigorously defined baseline:} A YOLO11s model trained with the architecture's default hyperparameters is adopted as the baseline, avoiding premature tuning that would confound the subsequent ablation analysis.
    \item \textbf{A systematic, one-factor-at-a-time ablation study:} Starting from the baseline, individual components --- augmentation policy, input resolution, optimizer configuration, and lightweight architectural enhancements inspired by prior attention-based apple-detection work \cite{Yang2024,Baek2025} --- are each varied in isolation to quantify their individual contribution to detection performance.
    \item \textbf{Design recommendations for controlled-environment deployment:} The resulting analysis yields practical guidance for real-time apple landmark detection systems operating under fixed-distance, fixed-illumination imaging conditions with high intra-class appearance variability.
\end{itemize}

\section{Related Work}
\label{sec:relwork}

\subsection{Deep Learning for Fruit Detection and the Evolution of YOLO}
\label{sec:relwork_dl}

Early fruit-detection research relied on image segmentation and classical computer-vision pipelines to localize fruit in orchard imagery and estimate yield \cite{Bargoti2017}, with related efforts extending to RGB-D-based in-field detection for other crops such as citrus \cite{Lin2019}. The subsequent shift toward single-stage deep detectors, beginning with the YOLO family, was motivated by the need to jointly satisfy accuracy and real-time inference constraints \cite{Bochkovskiy2020}. YOLOv4 in particular established an architecture-level balance between speed and accuracy that was widely adopted as a backbone for downstream agricultural detection tasks \cite{Bochkovskiy2020}, and related single-stage detectors were concurrently adapted to other fruit types, such as tomato \cite{Liu2020}, demonstrating the generality of the YOLO paradigm beyond apples.

\subsection{YOLO-Based Apple Detection}
\label{sec:relwork_yolo}

A substantial line of work has adapted successive YOLO versions specifically to apple detection under varying acquisition conditions. Tian et al. improved YOLOv3 for apple detection across different growth stages in orchards, addressing appearance changes over the growing season \cite{Tian2019}. Kang and Chen combined detection and segmentation for apple harvesting using a visual sensor mounted on a harvesting platform \cite{Kang2019}. Yan et al. proposed a real-time apple-target detection method based on an improved YOLOv5 for picking-robot applications \cite{Yan2021}, while Wang and He introduced a channel-pruned YOLOv5s model to accelerate apple fruitlet detection prior to fruit thinning \cite{Wang2021}. Zhang et al. developed a lightweight YOLOv4 variant to maintain detection accuracy in complex orchard backgrounds \cite{Zhang2022}, and Ji et al. combined ShuffleNetV2 with YOLOX to achieve a favorable speed--accuracy trade-off for picking-robot deployment \cite{Ji2022}. More recently, Yang et al. proposed CA-YOLOv5, which incorporates coordinate attention to improve apple detection in natural, unconstrained environments \cite{Yang2024}, Olgu\'in-Rojas et al. presented a lightweight YOLO-based architecture targeting embedded-system deployment \cite{OlguinRojas2025,Du2025}. Comprehensive surveys of the YOLO family \cite{Alif2024} and of deep learning applications across the apple production cycle \cite{Lv2025} situate these contributions within a broader trend toward progressively lighter and faster architectures for agricultural deployment.

\subsection{Landmark-Level and Attention-Based Apple Detection}
\label{sec:relwork_landmark}

While the studies above primarily address whole-fruit bounding-box detection, fewer works target sub-fruit morphological landmarks. The most directly related study is AS-YOLO, in which Baek et al. combined a Ghost Bottleneck module with a global attention mechanism to improve apple stem-region segmentation \cite{Baek2025}. This confirms that fine-grained landmarks such as the stem region benefit from architectural modifications beyond the detection backbones used for whole-fruit localization, a finding consistent with the coordinate-attention enhancement explored for whole-apple detection in natural environments \cite{Yang2024}. However, neither study isolates the individual contribution of each architectural or training-level modification through a controlled ablation procedure.

\subsection{Ablation Studies and Research Gap}
\label{sec:relwork_gap}

Across the reviewed literature, the predominant reporting practice is to present a single, already-optimized model configuration --- whether a pruned backbone \cite{Wang2021}, a lightweight variant \cite{Zhang2022,OlguinRojas2025,Du2025}, or an attention-augmented architecture \cite{Yang2024,Baek2025} --- without a systematic, controlled comparison of the individual design choices that produced it. Consequently, it remains unclear, particularly for small and fine-grained landmark classes such as the stem cavity and calyx, which individual factors (augmentation policy, input resolution, optimizer configuration, or lightweight architectural modules) drive the reported performance gains, and whether these effects generalize to imaging conditions with fixed camera distance and controlled illumination but high intra-class color and texture variability. The present study addresses this gap through a one-factor-at-a-time ablation study built on a rigorously defined YOLO11s baseline.

\section{Materials and Methods}
\label{sec:methods}

\subsection{Dataset Acquisition and Annotation}
\label{sec:dataset}

Unlike the majority of the reviewed studies, which rely on public, synthetic, or orchard-captured datasets \cite{Bargoti2017,Lin2019,Kang2019}, the dataset used in this study was collected entirely by the authors and is not derived from any existing public source. Images were captured under controlled, indoor conditions with a fixed camera-to-fruit distance and consistent illumination, eliminating variability due to ambient lighting or camera perspective. Within this controlled acquisition setup, the apples themselves exhibit substantial natural appearance variability, including bicoloration, surface blemishes, and a range of cultivar-dependent hues from deep red to pale yellow-green, which the subsequent ablation study is designed to account for.

All images were manually annotated by the authors using the Roboflow platform and exported in YOLO format, with three target classes: fruit body, stem cavity and calyx. The annotated dataset comprises 1506 images in total, partitioned into 1054 training, 301 validation, and 151 test images.

\subsection{Baseline Model Configuration}
\label{sec:baseline}

Following the trend toward lightweight YOLO variants for real-time agricultural deployment \cite{Wang2021,Zhang2022,Ji2022,OlguinRojas2025,Du2025}, the small variant of the YOLOv11 architecture (YOLO11s) is adopted as the baseline detector. To ensure that the subsequent ablation study isolates the effect of each individually varied factor, the baseline is trained using the architecture's default hyperparameters without prior tuning: an input resolution of $640 \times 640$ pixels, the framework's default optimizer and learning-rate schedule, and the default augmentation policy. Training is run for up to 100 epochs, consistent with established practice for datasets of comparable scale \cite{Bochkovskiy2020,Yan2021}, with early stopping applied via a validation-performance patience criterion (patience = 30 epochs without improvement). This patience value is held fixed across the baseline and every subsequent ablation run; it is a training-control setting rather than an ablated factor, so fixing it does not compromise the one-factor-at-a-time design. This untuned configuration serves as the fixed reference point against which every subsequent ablation variant is compared.

\subsection{Ablation Study Design}
\label{sec:ablation}

To decouple confounding interactions and establish the precise marginal utility of each engineering intervention, the ablation experiments follow a strict One-Factor-at-a-Time (OFAT) protocol. In complex computer vision pipelines, multi-parameter grid searches or simultaneous feature stacking often obscure whether performance improvements stem from genuine representational enhancements or accidental regularizing cross-cancellations. By modifying exactly one operational variable per experiment relative to the baseline configuration (Section~\ref{sec:baseline}) while clamping all remaining parameters to their default states, performance variations can be attributed unambiguously to that isolated intervention. This controlled experimental methodology directly addresses the methodological limitation identified in Section~\ref{sec:relwork_gap}, where prior agricultural vision studies predominantly present composite, post-hoc optimized models without isolating the causal trajectory of each individual component \cite{Wang2021,Zhang2022,Yang2024,OlguinRojas2025,Baek2025,Du2025}.

The experimental dimensions were selected to systematically probe the trade-offs between spatial representation, photometric robustness, convergence dynamics, and computational efficiency in high-speed sorting machinery:
\begin{enumerate}
    \item \textbf{Spatial Resolution Scaling:} Small anatomical landmarks occupy only a minor fraction of the total image area. Higher spatial resolutions increase sampling density across fine morphological contours (e.g., the concentric creasing of the stem cavity), but scale computational complexity quadratically and may introduce spatial quantization noise in self-attention layers.
    \item \textbf{Photometric and Geometric Augmentations:} In factory grading lines operating with fixed-focus industrial optics and constant LED strobes, geometric viewpoint variance is mechanically constrained. However, biological variations (e.g., anthocyanin blush gradients, russeting, and chlorophyll fading across cultivars) represent extreme photometric diversity, raising the question of whether synthetic color perturbations aid or corrupt subtle structural landmarks.
    \item \textbf{Optimization Trajectory:} Adaptive gradient methods interact sensitively with small-to-medium agricultural datasets. Evaluating hand-crafted weight decay schedules against automated framework heuristics determines whether convergence stability can be decoupled from empirical overparameterization.
    \item \textbf{Direction-Aware Structural Attention:} Morphological depressions possess pronounced directional symmetries. Evaluating whether coordinate-aware attention modules can outperform isotropic self-attention blocks without computational inflation addresses the core embedded latency requirement.
    \item \textbf{Post-Hoc Decision Thresholding:} Actuator triggering in industrial mechanical sorters operates under asymmetric cost matrices, where landmark false negatives lead directly to orientational disorientation and line recirculation.
\end{enumerate}

The five investigated factors, their specific parametric alterations, and their operational hypotheses are summarized in Table~\ref{tab:ablation}.

\begin{table}[!htbp]
\caption{Ablation factors, tested variants, and engineering rationale.}
\label{tab:ablation}
\small
\renewcommand{\arraystretch}{0.88}
\setlength{\tabcolsep}{3pt}
\begin{tabularx}{\textwidth}{c l X X}
\toprule
\textbf{\#} & \textbf{Factor} & \textbf{Variants Tested} & \textbf{Rationale} \\
\midrule
1 & Input resolution & $640\times640$; $768\times768$ & High resolution may improve small landmark sampling but incurs quadratic compute scaling. \\
2 & Augmentation policy & CutMix; Mixup; HSV jitter; Rotation ($15^\circ$) & Fixed industrial geometry versus biological surface hue and texture variability across cultivars. \\
3 & Optimizer / LR & Default (auto) vs.\ AdamW + cosine LR & Assesses convergence stability and risk of overfitting on a clean, moderately sized dataset. \\
4 & Architectural modules & Same-slot Coordinate Attention (CA) swap & Position-sensitive channel attention replaces isotropic self-attention (C2PSA) for landmarks. \\
5 & Decision threshold & Post-hoc confidence sweep ($0.10 \le \tau \le 0.70$) & Balances precision and recall against asymmetric costs of sorting actuator false negatives. \\
\bottomrule
\end{tabularx}

\vspace{4pt}
\normalsize
\noindent\textbf{Experimental Rationale.} The structured breakdown in Table~\ref{tab:ablation} establishes an isolated empirical progression across all critical phases of the visual pipeline. By sequentially interrogating data representation (Factor~1), data manifold regularization (Factor~2), loss-surface optimization (Factor~3), structural inductive bias (Factor~4), and deployment operating semantics (Factor~5), each component's distinct contribution to fine-grained sub-fruit localization is explicitly mapped without conflation.
\end{table}

\subsection{Hardware and Software Infrastructure}
\label{sec:hardware}

The baseline model and all subsequent ablation variants were trained and evaluated on a workstation equipped with a 13th Gen Intel Core i7-13700HX CPU and an NVIDIA GeForce RTX 5060 Laptop GPU. Training was implemented in Python using the PyTorch deep learning framework and the Ultralytics YOLO11 implementation.

\section{Results}
\label{sec:results}

\subsection{Baseline Model Performance}
\label{sec:baseline_results}

Training of the baseline YOLO11s model converged before reaching the 100-epoch budget: early stopping (patience = 30) triggered at epoch 88, with the best checkpoint retained from epoch 58. Table~\ref{tab:baseline} reports precision, recall, mAP@0.5, and mAP@0.5:0.95 for each class on the held-out test set (151 images, 233 annotated instances).

\begin{table}[!htbp]
\caption{Baseline performance on the held-out test split.}
\label{tab:baseline}
\small
\renewcommand{\arraystretch}{0.88}
\setlength{\tabcolsep}{3.5pt}
\begin{tabularx}{\textwidth}{CCCCCC}
\toprule
\textbf{Class} & \textbf{Precision} & \textbf{Recall} & \textbf{mAP@0.5} & \textbf{mAP@0.5:0.95} & \textbf{F1} \\
\midrule
Fruit body & 0.980 & 1.000 & \textbf{0.992} & 0.991 & 0.9901 \\
Calyx & 0.909 & 0.923 & \textbf{0.918} & 0.507 & 0.9160 \\
Stem cavity & 0.672 & 0.963 & \textbf{0.853} & 0.382 & 0.7918 \\
\midrule
\textbf{All classes} & \textbf{0.854} & \textbf{0.962} & \textbf{0.921} & \textbf{0.627} & \textbf{0.9047} \\
\bottomrule
\end{tabularx}

\vspace{4pt}
\normalsize
\noindent\textbf{Baseline Analysis.} As indicated in Table~\ref{tab:baseline}, the macro-level fruit body is localized with near-perfect reliability (F1 = 0.9901), indicating that whole-object bounding is virtually solved by modern single-stage backbones under controlled lighting. Conversely, a pronounced performance disparity is evident among the morphological sub-landmarks: while the calyx achieves a strong F1 of 0.9160, the stem cavity demonstrates substantially depressed precision (0.672) and tight IoU precision (mAP@0.5:0.95 = 0.382). This confirms that visual ambiguity, shallow depth-of-field shading, and skin-tone transitions predominantly degrade the smallest anatomical depression, underscoring the stem cavity as the primary bottleneck for orientation alignment.
\end{table}

\subsection{Ablation Factor 1 --- Input Resolution}
\label{sec:resolution}

The $768\times768$ variant was trained with the same baseline hyperparameters (patience = 30), converging via early stopping at epoch 95 with the best checkpoint retained from epoch 65. Table~\ref{tab:res768} reports per-class test-set performance.

\begin{table}[!htbp]
\caption{Input-resolution variant ($768\times768$) performance on the test set.}
\label{tab:res768}
\centering
\begin{tabular}{cccccc}
\toprule
\textbf{Class} & \textbf{Precision} & \textbf{Recall} & \textbf{mAP@0.5} & \textbf{mAP@0.5:0.95} & \textbf{F1} \\
\midrule
Fruit body & 0.983 & 1.000 & \textbf{0.985} & 0.985 & 0.9905 \\
Calyx & 0.963 & 0.808 & \textbf{0.944} & 0.489 & 0.9301 \\
Stem cavity & 0.827 & 0.745 & \textbf{0.790} & 0.377 & 0.7382 \\
\midrule
\textbf{All classes} & \textbf{0.924} & \textbf{0.851} & \textbf{0.907} & \textbf{0.617} & \textbf{0.8896} \\
\bottomrule
\end{tabular}
\end{table}

Contrary to the hypothesis in Section~\ref{sec:ablation} that a higher input resolution would improve detection of the small stem-cavity and calyx landmarks, increasing the resolution from $640\times640$ to $768\times768$ degrades overall performance relative to the baseline: mAP@0.5 drops from 0.921 to 0.907, mAP@0.5:0.95 from 0.627 to 0.617, and the overall F1-score from 0.9047 to 0.8896. The effect is strongest for the stem cavity class, whose recall falls from 0.963 to 0.745 and mAP@0.5 from 0.853 to 0.790. Precision increases slightly overall (0.854 to 0.924), indicating the higher-resolution model becomes more conservative, missing more true instances rather than producing more false positives. Inference speed is also substantially affected, with throughput nearly halving (142.8 to 141.0 FPS; 7.00 to 7.09~ms per image). Based on this result, $640\times640$ is retained as the preferred resolution.

\subsection{Ablation Factor 2 --- Augmentation Policy}
\label{sec:augmentation}

With the input resolution fixed at $640\times640$, this factor evaluates whether modifying the default photometric, geometric, or blend-based augmentation parameters improves landmark detection across varying apple cultivars.

\begin{table}[!htbp]
\caption{Ablation Factor 2, Run E variant tested.}
\label{tab:aug-variants}
\centering
\begin{tabular}{p{2.6cm}p{4.2cm}p{4.6cm}}
\toprule
\textbf{Run} & \textbf{Changed parameter(s)} & \textbf{Default $\rightarrow$ Tested} \\
\midrule
E --- Stronger HSV & \texttt{hsv\_s}, \texttt{hsv\_v} & 0.7 / 0.4 $\rightarrow$ 0.9 / 0.6 \\
\bottomrule
\end{tabular}
\end{table}

Run E increases HSV saturation and value jitter beyond the default, testing whether a stronger synthetic color perturbation can further improve robustness to the dataset's natural inter-cultivar color variation (Table~\ref{tab:aug-variants}). Vertical-flip augmentation was excluded because vertical flipping would produce physically implausible fruit orientations for a task centered on detecting orientation itself.

\paragraph{Run E --- Stronger HSV.}
Training converged via early stopping at epoch 85, with the best checkpoint retained from epoch 55.

\begin{table}[!htbp]
\caption{Run E (stronger HSV) performance on the test set.}
\label{tab:augE}
\centering
\begin{tabular}{cccccc}
\toprule
\textbf{Class} & \textbf{Precision} & \textbf{Recall} & \textbf{mAP@0.5} & \textbf{mAP@0.5:0.95} & \textbf{F1} \\
\midrule
Fruit body & 0.983 & 1.000 & \textbf{0.990} & 0.990 & 0.9915 \\
Calyx & 0.938 & 0.876 & \textbf{0.964} & 0.501 & 0.9062 \\
Stem cavity & 0.747 & 0.812 & \textbf{0.811} & 0.330 & 0.7781 \\
\midrule
\textbf{All classes} & \textbf{0.889} & \textbf{0.896} & \textbf{0.922} & \textbf{0.607} & \textbf{0.8928} \\
\bottomrule
\end{tabular}
\end{table}

Run E produces a mixed result relative to the baseline (Table~\ref{tab:augE}). Overall mAP@0.5 edges marginally higher (0.922 vs.\ 0.921), driven almost entirely by a calyx-class improvement (mAP@0.5 of 0.964 vs.\ 0.918, the best calyx result of any variant tested so far). However, every other aggregate metric is worse than the baseline: mAP@0.5:0.95 falls from 0.627 to 0.607, recall from 0.962 to 0.896, and overall F1 from 0.9047 to 0.8928. The stem cavity class in particular regresses on mAP@0.5:0.95 (0.382 to 0.330) despite a small recall gain.

\paragraph{Run F --- Rotation augmentation.}
Random image rotation is left at $0^\circ$ (disabled) in the Ultralytics default and in every run above. Since the apple's orientation relative to the stem/calyx axis is exactly what the downstream alignment system must determine, introducing rotation during training may help the model learn landmark appearance independent of in-plane orientation, rather than relying on an implicit ``upright'' prior from the fixed camera setup (Table~\ref{tab:augF-variant}).

\begin{table}[!htbp]
\caption{Ablation Factor 2, Run F variant tested.}
\label{tab:augF-variant}
\centering
\begin{tabular}{p{2.6cm}p{4.2cm}p{4.6cm}}
\toprule
\textbf{Run} & \textbf{Changed parameter(s)} & \textbf{Default $\rightarrow$ Tested} \\
\midrule
F --- Rotation on & \texttt{degrees} & 0.0 $\rightarrow$ 15.0 \\
\bottomrule
\end{tabular}
\end{table}

\paragraph{Run F results.}
Training converged via early stopping at epoch 98, with the best checkpoint retained from epoch 68.

\begin{table}[!htbp]
\caption{Run F (rotation on) performance on the test set.}
\label{tab:augF}
\centering
\begin{tabular}{cccccc}
\toprule
\textbf{Class} & \textbf{Precision} & \textbf{Recall} & \textbf{mAP@0.5} & \textbf{mAP@0.5:0.95} & \textbf{F1} \\
\midrule
Fruit body & 0.982 & 1.000 & \textbf{0.994} & 0.987 & 0.9908 \\
Calyx & 0.959 & 0.893 & \textbf{0.916} & 0.497 & 0.9248 \\
Stem cavity & 0.629 & 0.906 & \textbf{0.827} & 0.386 & 0.7428 \\
\midrule
\textbf{All classes} & \textbf{0.857} & \textbf{0.933} & \textbf{0.912} & \textbf{0.623} & \textbf{0.8932} \\
\bottomrule
\end{tabular}
\end{table}

Adding rotation augmentation does not exceed the baseline either (Table~\ref{tab:augF}): overall mAP@0.5 (0.912 vs.\ 0.921), mAP@0.5:0.95 (0.623 vs.\ 0.627), and F1 (0.8932 vs.\ 0.9047) are all slightly lower. The pattern resembles earlier non-default-exceeding runs --- a modest calyx-class improvement (F1 0.9248, the best calyx F1 of any run so far, edging out even Run E) paired with a stem-cavity precision drop (0.629 vs.\ 0.672 baseline), suggesting rotation introduces more false positives for the smallest, most ambiguous class even as it helps the model generalize calyx appearance.

\paragraph{Run G results (Mixup, 0.1).}
Training converged via early stopping at epoch 90, with the best checkpoint retained from epoch 60.

\begin{table}[!htbp]
\caption{Run G (mixup on) performance on the test set.}
\label{tab:augG}
\centering
\begin{tabular}{cccccc}
\toprule
\textbf{Class} & \textbf{Precision} & \textbf{Recall} & \textbf{mAP@0.5} & \textbf{mAP@0.5:0.95} & \textbf{F1} \\
\midrule
Fruit body & 0.984 & 1.000 & \textbf{0.990} & 0.990 & 0.9918 \\
Calyx & 0.948 & 0.904 & \textbf{0.937} & 0.480 & 0.9253 \\
Stem cavity & 0.795 & 0.726 & \textbf{0.833} & 0.341 & 0.7586 \\
\midrule
\textbf{All classes} & \textbf{0.909} & \textbf{0.877} & \textbf{0.920} & \textbf{0.604} & \textbf{0.8923} \\
\bottomrule
\end{tabular}
\end{table}

Mixup at a conservative blend strength neither confirms nor refutes the original blurring concern cleanly (Table~\ref{tab:augG}): precision rises to the highest of any run so far (0.909, exceeding the baseline's 0.854), but recall drops accordingly (0.877 vs.\ 0.962), and mAP@0.5:0.95 falls to 0.604, among the lowest recorded. Overall F1 (0.8923) remains below the baseline (0.9047). The precision/recall trade-off is consistent with the blurring hypothesis: the model appears to become more conservative and miss more true instances, rather than localizing them less accurately, when trained on blended images.

\paragraph{Run H results (CutMix, 0.1).}
Training ran the full 100-epoch budget without early stopping triggering.

\begin{table}[!htbp]
\caption{Run H (CutMix on) performance on the test set.}
\label{tab:augH}
\centering
\begin{tabular}{cccccc}
\toprule
\textbf{Class} & \textbf{Precision} & \textbf{Recall} & \textbf{mAP@0.5} & \textbf{mAP@0.5:0.95} & \textbf{F1} \\
\midrule
Fruit body & 0.983 & 1.000 & \textbf{0.990} & 0.989 & 0.9915 \\
Calyx & 0.920 & 0.904 & \textbf{0.925} & 0.498 & 0.9117 \\
Stem cavity & 0.782 & 0.719 & \textbf{0.756} & 0.350 & 0.7492 \\
\midrule
\textbf{All classes} & \textbf{0.895} & \textbf{0.874} & \textbf{0.890} & \textbf{0.612} & \textbf{0.8845} \\
\bottomrule
\end{tabular}
\end{table}

CutMix produces the weakest result of the two blend-based augmentations (Table~\ref{tab:augH}): mAP@0.5 falls to 0.890 (the lowest of any run tested), and F1 (0.8845) also trails both the baseline and Run G. Unlike mixup, CutMix does not raise precision noticeably (0.895 vs.\ baseline 0.854) while still losing substantial recall (0.874 vs.\ 0.962), indicating the hard, rectangular region replacement used by CutMix is more disruptive to this task than mixup's soft blending --- plausibly because it can fully occlude a small landmark (stem cavity or calyx) rather than merely blending its appearance.

\paragraph{Summary of Ablation Factor 2 (Augmentation Policy) --- complete.}
Six variants (Runs E--H, plus two earlier removal tests, both superseded) were tested against the baseline; none exceeded it on the primary metrics (F1, mAP@0.5:0.95). Results are ranked in Table~\ref{tab:aug-summary} by overall F1:

\begin{table}[!htbp]
\caption{Ablation Factor 2 summary: all augmentation variants ranked by F1.}
\label{tab:aug-summary}
\centering
\begin{tabular}{cccc}
\toprule
\textbf{Configuration} & \textbf{F1} & \textbf{mAP@0.5} & \textbf{mAP@0.5:0.95} \\
\midrule
\textbf{Baseline (default augmentation)} & \textbf{0.9047} & \textbf{0.921} & \textbf{0.627} \\
Run F --- Rotation ($15^\circ$) & 0.8932 & 0.912 & 0.623 \\
Run E --- Stronger HSV & 0.8928 & 0.922 & 0.607 \\
Run G --- Mixup (0.1) & 0.8923 & 0.920 & 0.604 \\
Run H --- CutMix (0.1) & 0.8845 & 0.890 & 0.612 \\
\bottomrule
\end{tabular}
\end{table}

The default Ultralytics augmentation policy is therefore retained as the final configuration for Ablation Factor 2, and carried forward unchanged into all subsequent ablation factors. While no single modification improved the aggregate metrics, several runs (E, F, G) produced the best calyx-class results observed in this study, suggesting that further gains for that specific class --- rather than the model as a whole --- may be achievable with a more targeted, class-weighted training strategy, a direction left for future work.

\subsection{Ablation Factor 3 --- Optimizer / LR Schedule}
\label{sec:optimizer}

With the augmentation policy fixed at the Ultralytics default (Section~\ref{sec:augmentation}), this factor tests whether a manually configured optimizer and learning-rate schedule outperform the framework's automatic selection. All other settings (augmentation, imgsz = 640, patience = 30) are held at the baseline configuration (Table~\ref{tab:optim-variant}).

\begin{table}[!htbp]
\caption{Ablation Factor 3 variant tested.}
\label{tab:optim-variant}
\centering
\begin{tabular}{p{2.6cm}p{4.2cm}p{4.6cm}}
\toprule
\textbf{Run} & \textbf{Changed parameter(s)} & \textbf{Default $\rightarrow$ Tested} \\
\midrule
I --- AdamW + cosine LR & \texttt{optimizer}, \texttt{lr0}, \texttt{cos\_lr} & auto / 0.01 / off $\rightarrow$ AdamW / 0.001 / on \\
\bottomrule
\end{tabular}
\end{table}

The baseline's \texttt{optimizer=auto} setting lets Ultralytics select between SGD and AdamW based on dataset size and epoch count, with a linear learning-rate decay. Run I instead forces AdamW with a lower initial learning rate (0.001, a common default for AdamW) and a cosine annealing schedule, testing whether this more commonly hand-tuned configuration converges to a better optimum than the automatic default.

\paragraph{Run I results (AdamW, lr0 = 0.001, cosine LR).}
Training ran the full 100-epoch budget without early stopping triggering.

\begin{table}[!htbp]
\caption{Run I (AdamW + cosine LR) performance on the test set.}
\label{tab:optimI}
\centering
\begin{tabular}{cccccc}
\toprule
\textbf{Class} & \textbf{Precision} & \textbf{Recall} & \textbf{mAP@0.5} & \textbf{mAP@0.5:0.95} & \textbf{F1} \\
\midrule
Fruit body & 0.982 & 1.000 & \textbf{0.984} & 0.984 & 0.9911 \\
Calyx & 0.936 & 0.865 & \textbf{0.922} & 0.513 & 0.8992 \\
Stem cavity & 0.713 & 0.875 & \textbf{0.856} & 0.320 & 0.7859 \\
\midrule
\textbf{All classes} & \textbf{0.877} & \textbf{0.913} & \textbf{0.921} & \textbf{0.606} & \textbf{0.8949} \\
\bottomrule
\end{tabular}
\end{table}

The manually configured AdamW optimizer with cosine learning-rate annealing matches the baseline almost exactly on mAP@0.5 (0.921 vs.\ 0.921) but does not exceed it on either of the primary metrics (Table~\ref{tab:optimI}): mAP@0.5:0.95 falls from 0.627 to 0.606, and F1 from 0.9047 to 0.8949. Precision improves modestly (0.877 vs.\ 0.854), but at the cost of recall (0.913 vs.\ 0.962), the same precision/recall trade-off pattern observed in several augmentation runs. The stem cavity class shows the sharpest mAP@0.5:0.95 decline (0.382 to 0.320), the weakest result for that metric on this class across all ablation factors tested to date.

\paragraph{Summary of Ablation Factor 3 (Optimizer / LR Schedule).}
Ultralytics' automatic optimizer selection (\texttt{optimizer=auto}) outperforms the manually configured AdamW + cosine-annealing alternative on every primary metric. The automatic default is therefore retained, and the baseline configuration remains the best-performing setup identified across all ablation factors examined so far (input resolution, augmentation policy, and optimizer/LR schedule).

\subsection{Ablation Factor 4 --- Lightweight Architectural Module}
\label{sec:ca_swap}

\begin{figure}[!htbp]
\centering
\includegraphics[width=0.84\textwidth]{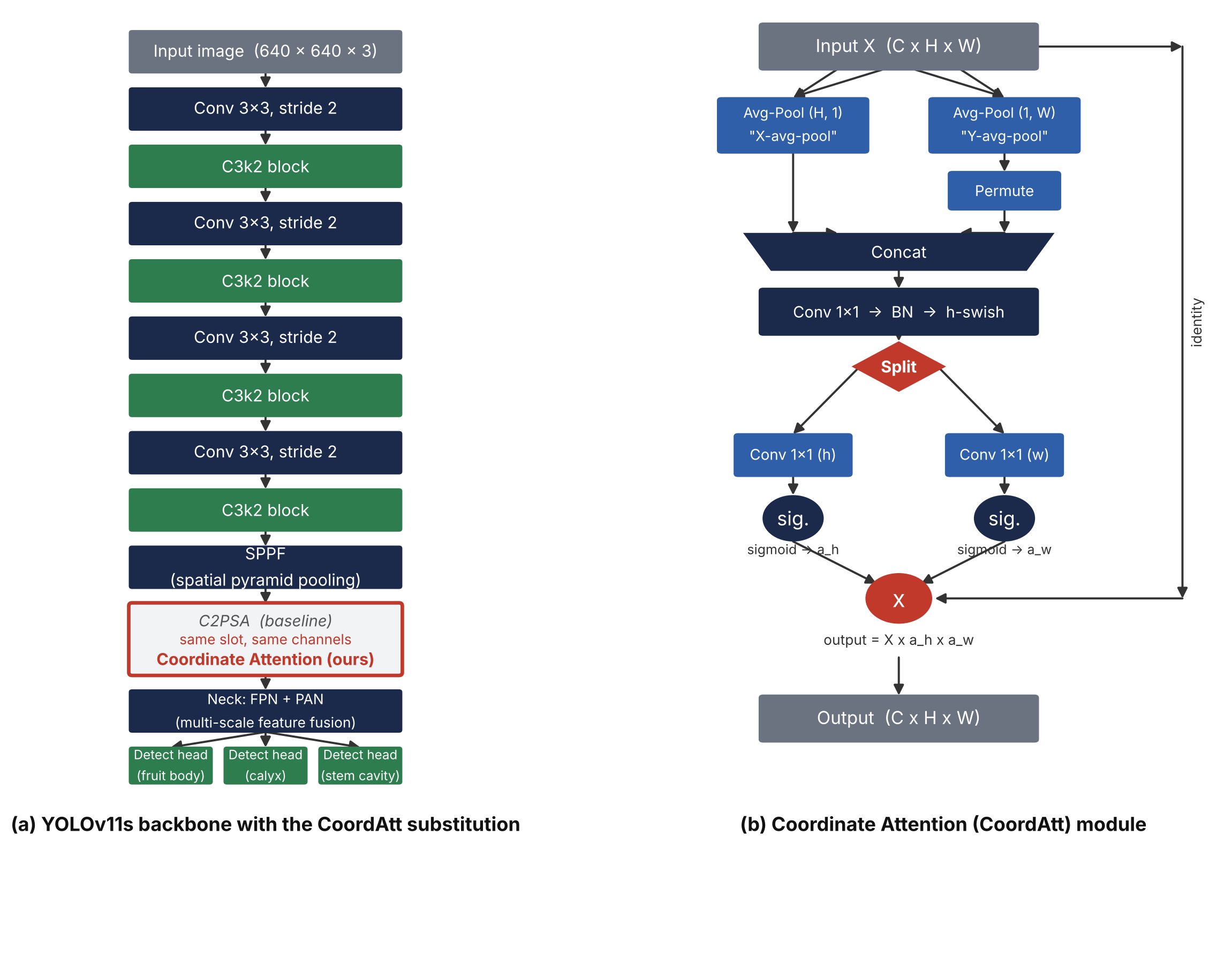}
\vspace{-0.9cm}
\caption{(\textbf{a}) YOLO11s backbone with the Coordinate Attention (CoordAtt) substitution: CoordAtt occupies the same architectural slot as the baseline's C2PSA block, with identical input/output channel counts, so no other layer in the backbone, neck, or detection heads is modified. (\textbf{b}) Internal dataflow of the CoordAtt module: the input feature map is pooled separately along the height and width axes, concatenated, passed through a shared $1\times1$ convolution--batch-norm--h-swish bottleneck, split back into two branches, and used to generate a pair of sigmoid attention maps ($a_h$, $a_w$) that rescale the original input via element-wise multiplication.}
\label{fig:architecture}
\end{figure}

Rather than stacking an additional attention mechanism onto YOLO11s's existing C2PSA block (which already performs spatial self-attention at the end of the backbone), this factor swaps C2PSA for Coordinate Attention (CA) in the same architectural slot --- same position, same input/output channel count --- following the design used in CA-YOLOv5 for apple detection \cite{Yang2024}. This same-slot substitution avoids both the index-renumbering risk of inserting a new layer and the redundancy risk of running two different attention mechanisms back to back, isolating the architectural comparison to attention \textit{type} rather than attention \textit{quantity}. All other settings (augmentation, imgsz~=~640, patience~=~30, default optimizer) match the baseline exactly. Figure~\ref{fig:architecture} illustrates (a) where this substitution occurs within the YOLO11s backbone and (b) the internal dataflow of the Coordinate Attention module itself.

\textbf{Results.} Training converged via early stopping at epoch 73, with the best checkpoint retained from epoch 43. The CA-swap model is also lighter than the baseline: 8,451,449 parameters and 20.5 GFLOPs, versus 9,413,961 parameters and 21.4 GFLOPs for the unmodified YOLO11s --- a 10.2\% parameter reduction and 4.2\% fewer GFLOPs, since Coordinate Attention is a cheaper mechanism than C2PSA (Table~\ref{tab:ca_swap}).

\begin{table}[!htbp]
\caption{Per-class test-set results for the Coordinate Attention (CA) swap (Ablation Factor 4).}
\label{tab:ca_swap}
\small
\renewcommand{\arraystretch}{0.90}
\setlength{\tabcolsep}{3.5pt}
\begin{tabularx}{\textwidth}{CCCCCC}
\toprule
\textbf{Class} & \textbf{Precision} & \textbf{Recall} & \textbf{mAP@0.5} & \textbf{mAP@0.5:0.95} & \textbf{F1} \\
\midrule
Fruit body & 0.982 & 1.000 & 0.984 & 0.984 & 0.9908 \\
Calyx & 0.903 & 0.891 & 0.949 & 0.494 & 0.8969 \\
Stem cavity & 0.777 & 0.872 & 0.870 & 0.342 & 0.8217 \\
\midrule
\textbf{All classes} & \textbf{0.887} & \textbf{0.921} & \textbf{0.935} & \textbf{0.607} & \textbf{0.9038} \\
\bottomrule
\end{tabularx}
\end{table}

\begin{figure}[!htbp]
\centering
\includegraphics[width=0.84\textwidth]{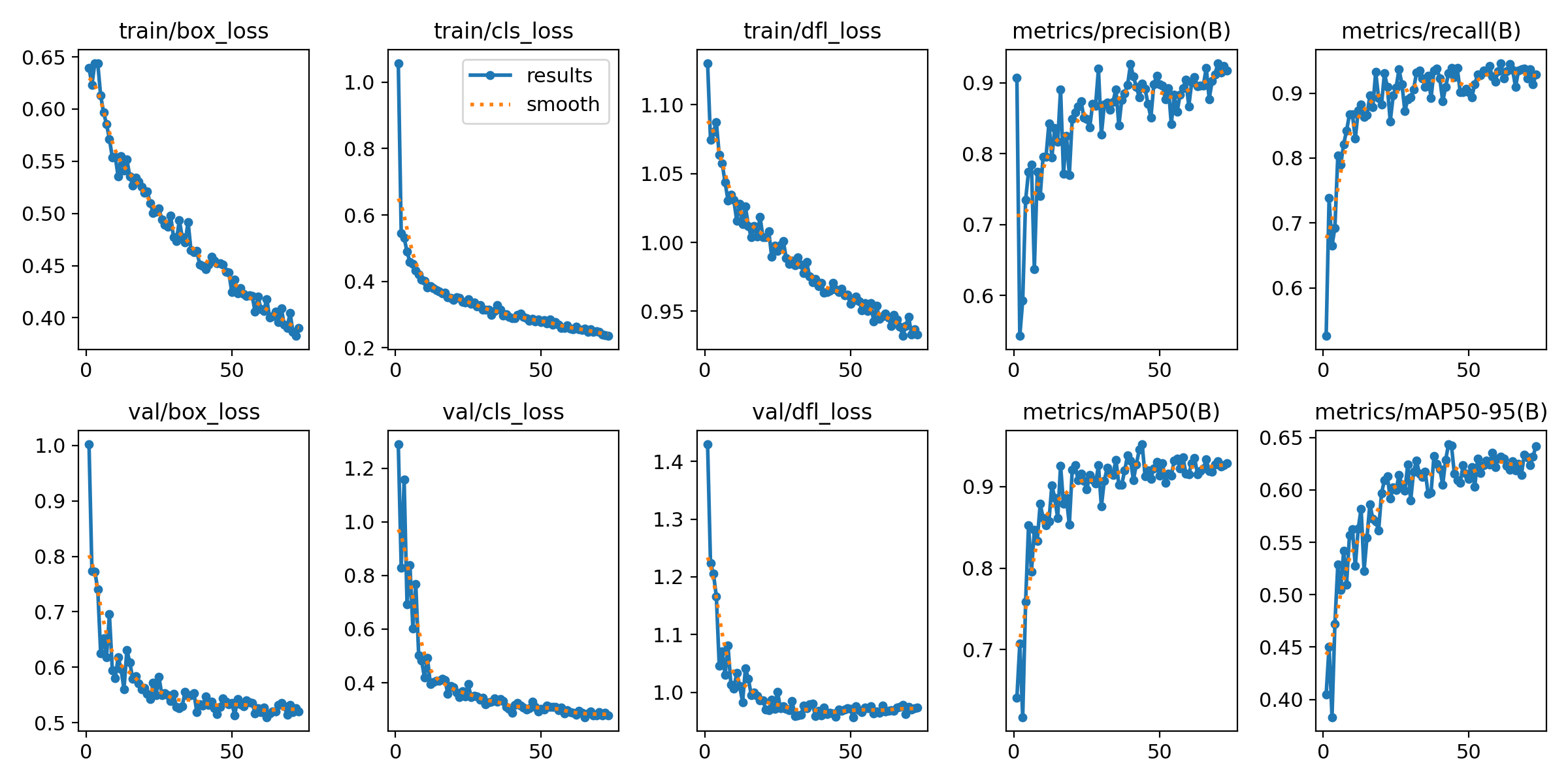}
\caption{Training and validation curves for the final CA-swap model across 73 training epochs (early stopping triggered at epoch 73, best checkpoint retained from epoch 43): box, classification, and DFL loss (train and validation), and validation-split precision, recall, mAP@0.5, and mAP@0.5:0.95.}
\label{fig:ca_training_curves}
\end{figure}

\begin{figure}[!htbp]
\centering
\includegraphics[width=0.72\textwidth]{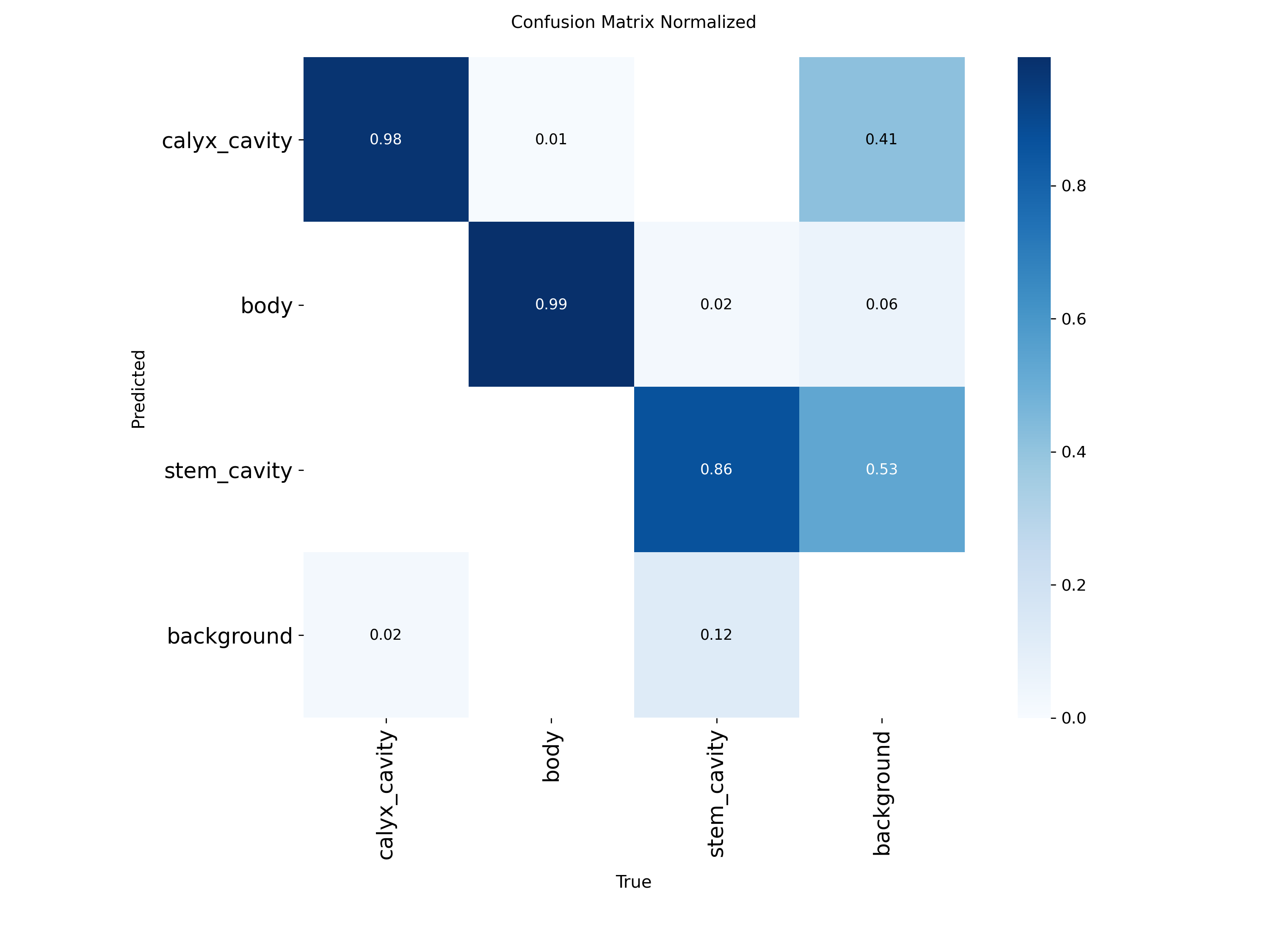}
\caption{Normalized confusion matrix for the final CA-swap model on the validation split (301 images), showing per-class prediction accuracy and the degree to which each class is confused with background.}
\label{fig:ca_confusion_matrix}
\end{figure}

\noindent\textbf{Discussion.} The CA-swap model achieves an overall F1 of 0.9038, within 0.1 percentage points of the baseline (0.9047) --- the closest any tested variant has come to matching the baseline across every ablation factor examined so far --- while using roughly 10\% fewer parameters. Overall mAP@0.5 is the highest recorded in this study (0.935 vs.\ baseline 0.921), driven by gains in both calyx (0.949 vs.\ 0.918) and stem cavity (0.870 vs.\ 0.853). Most notably, the stem cavity class --- the weakest class in every prior configuration --- achieves its best F1 of the entire study (0.8217, driven by a large precision gain from 0.672 to 0.777), suggesting that coordinate attention's position-sensitive channel weighting is specifically well suited to localizing this small, fine-grained, variably textured landmark. The trade-off is a lower mAP@0.5:0.95 (0.607 vs.\ 0.627) and a modest calyx F1 decline (0.8969 vs.\ 0.9160), indicating the swap sharpens coarse localization (mAP@0.5) and the hardest class (stem cavity) at some cost to the tightest bounding-box precision (mAP@0.5:0.95) and the already-strong calyx class. Training diagnostics and confusion matrix behavior for this model are shown in Figures~\ref{fig:ca_training_curves} and \ref{fig:ca_confusion_matrix}.

\noindent\textbf{Summary of Ablation Factor 4.} Replacing C2PSA with Coordinate Attention is the strongest result identified across all ablation factors tested (resolution, augmentation, optimizer/LR, and architecture): it matches the baseline's overall F1 almost exactly, improves mAP@0.5 and stem-cavity detection specifically, and reduces model size and compute --- a meaningful advantage for the real-time, embedded deployment scenario motivating this study. This configuration is carried forward as a candidate for the final combined model, alongside the retained default augmentation policy (Section~\ref{sec:augmentation}) and default optimizer (Section~\ref{sec:optimizer}).

\subsection{Ablation Factor 5 --- Inference Decision Threshold}
\label{sec:threshold}

This factor evaluates post-hoc confidence thresholding on the CA-swap model (Section~\ref{sec:ca_swap}) without retraining, holding the NMS IoU threshold at its default (0.7) across 0.10 to 0.70 on the held-out test split (Table~\ref{tab:conf_sweep}).

\begin{table}[!htbp]
\caption{Confidence-threshold sweep on the CA-swap model's test-set predictions (Ablation Factor 5).}
\label{tab:conf_sweep}
\small
\renewcommand{\arraystretch}{0.90}
\setlength{\tabcolsep}{3pt}
\begin{tabularx}{\textwidth}{CCCCCC}
\toprule
\textbf{Conf.\ threshold} & \textbf{Precision} & \textbf{Recall} & \textbf{F1} & \textbf{mAP@0.5} & \textbf{mAP@0.5:0.95} \\
\midrule
0.10 & 0.8885 & 0.9189 & 0.9035 & 0.9108 & 0.5959 \\
0.15 & 0.8888 & 0.9205 & \textbf{0.9043} & \textbf{0.9108} & 0.5959 \\
0.20 & 0.8888 & 0.9199 & 0.9041 & 0.8900 & 0.5875 \\
0.25 & 0.8967 & 0.8990 & 0.8979 & 0.8743 & 0.5802 \\
0.30 & 0.9106 & 0.8882 & 0.8992 & 0.8666 & 0.5764 \\
0.35 & 0.9107 & 0.8886 & 0.8995 & 0.8666 & 0.5764 \\
0.40 & 0.9085 & 0.8778 & 0.8929 & 0.8559 & 0.5729 \\
0.50 & 0.9458 & 0.8405 & 0.8901 & 0.8238 & 0.5634 \\
0.60 & 0.9629 & 0.7989 & 0.8732 & 0.7860 & 0.5478 \\
0.70 & 0.9465 & 0.6883 & 0.7970 & 0.6787 & 0.5022 \\
\bottomrule
\end{tabularx}

\vspace{6pt}
\normalsize
\noindent\textbf{Discussion.} F1 peaks at a confidence threshold of 0.15 (P~=~0.8888, R~=~0.9205, F1~=~0.9043), which also ties the highest mAP@0.5 (0.9108) recorded in the sweep. As the threshold rises, precision climbs steadily (reaching 0.9629 at 0.60) while recall falls sharply (to 0.7989 at the same point and 0.6883 at 0.70), and F1 declines monotonically past 0.15 --- confirming that, for this model, low-to-moderate thresholds dominate the precision--recall trade-off rather than Ultralytics' conventional default of 0.25, which already underperforms (F1~=~0.8979) relative to 0.10--0.20 (Figure~\ref{fig:ca_pr_f1_curves}). Because missed stem-cavity or calyx detections directly translate into failed orientation decisions downstream in the sorting pipeline, the cost of a false negative is higher than that of a false positive, which further favors operating near the recall-leaning end of the curve rather than strictly at the F1-maximizing point.
\end{table}

\begin{figure}[!htbp]
\centering
\begin{minipage}[b]{0.48\textwidth}
    \centering
    \includegraphics[width=\textwidth]{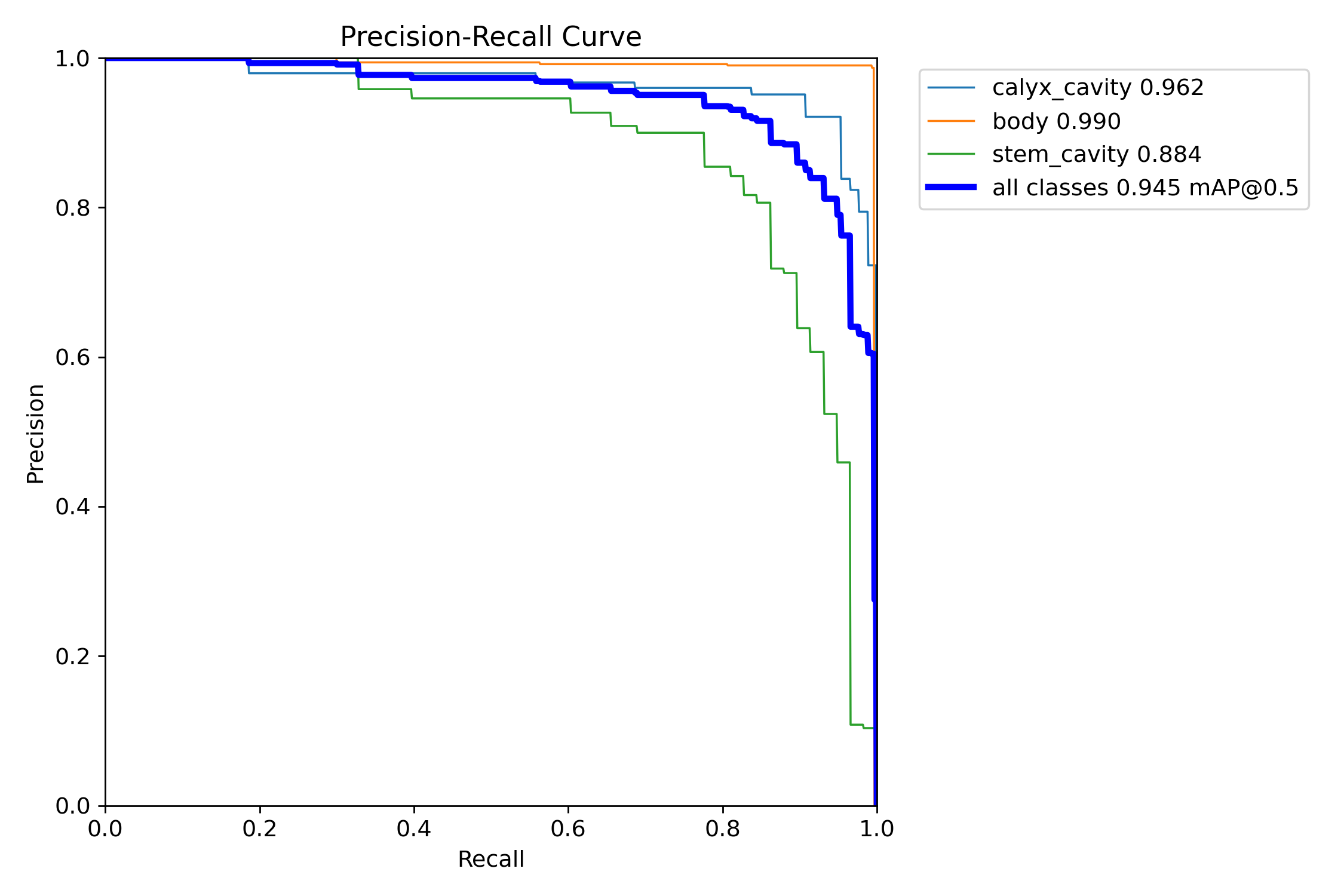}\\[3pt]
    {\normalsize \textbf{(a)}}
\end{minipage}
\hfill
\begin{minipage}[b]{0.48\textwidth}
    \centering
    \includegraphics[width=\textwidth]{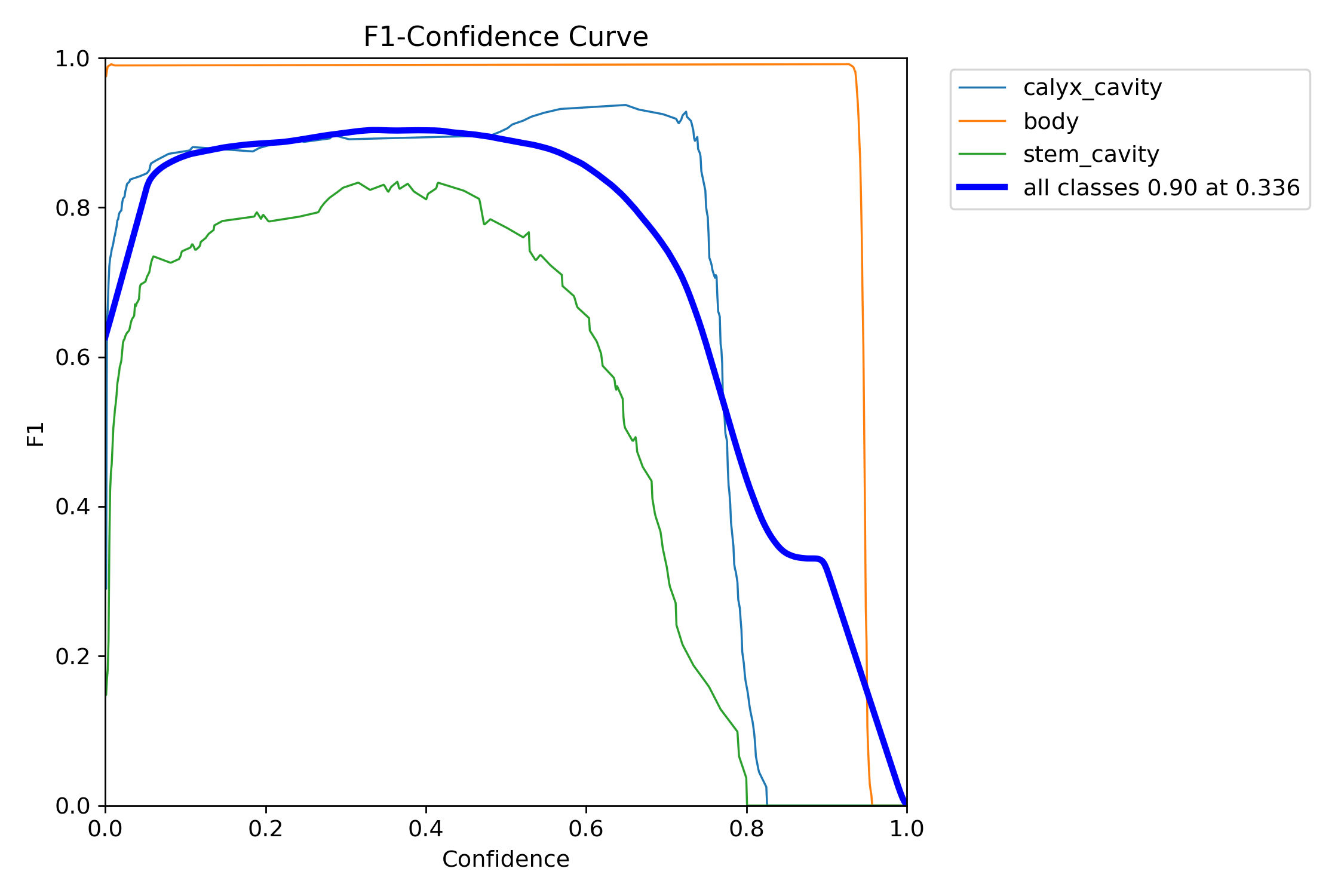}\\[3pt]
    {\normalsize \textbf{(b)}}
\end{minipage}
\caption{(\textbf{a}) Precision--recall curve and (\textbf{b}) F1-confidence curve for the final CA-swap model, computed on the validation split (301 images) as a training-time diagnostic. Note: the per-point optimum shown in (\textbf{b}) (confidence $\approx 0.336$) reflects the validation split and is not the deployment recommendation; the 0.15 confidence threshold adopted in this study was instead selected via an explicit sweep on the held-out test split (Table~\ref{tab:conf_sweep}), the split used for every other reported result in this paper.}
\label{fig:ca_pr_f1_curves}
\end{figure}

\textbf{Summary of Ablation Factor 5.} A confidence threshold of 0.15 is selected as the deployment operating point for the CA-swap model: it maximizes F1, matches the sweep's best mAP@0.5, and sits at the recall-favorable end of the precision--recall trade-off appropriate for an orientation system where missed landmark detections are more costly than occasional false positives. This closes the five-factor ablation study; the final recommended configuration combines the default augmentation policy (Section~\ref{sec:augmentation}), the default (automatic) optimizer/LR schedule (Section~\ref{sec:optimizer}), the CoordAtt architectural swap (Section~\ref{sec:ca_swap}), and this 0.15 confidence threshold (Section~\ref{sec:threshold}).

\subsection{Final Combined Model --- Testing Additivity}
\label{sec:final_combined}

Sections~\ref{sec:augmentation} and \ref{sec:ca_swap} identified two variants that each improved on the baseline's calyx-class performance in isolation: Run E (stronger HSV augmentation) produced the best calyx mAP@0.5 of any augmentation variant (0.964), while the CA-swap model (Section~\ref{sec:ca_swap}) was the single strongest overall configuration but showed a modest calyx F1 decline relative to the baseline (0.8969 vs.\ 0.9160). Since ablation effects are not guaranteed to compose additively, this was tested directly rather than assumed: a final combined model was trained using the CA-swap architecture together with Run E's stronger HSV settings (hsv\_s~=~0.9, hsv\_v~=~0.6), with every other setting --- imgsz~=~640, patience~=~30, default optimizer --- unchanged (Table~\ref{tab:final_combined}).

\begin{table}[!htbp]
\caption{Per-class test-set results for the final combined model (CA-swap architecture + stronger HSV augmentation, Section~\ref{sec:final_combined}).}
\label{tab:final_combined}
\begin{tabularx}{\textwidth}{CCCCCC}
\toprule
\textbf{Class} & \textbf{Precision} & \textbf{Recall} & \textbf{mAP@0.5} & \textbf{mAP@0.5:0.95} & \textbf{F1} \\
\midrule
Fruit body & 0.984 & 1.000 & 0.994 & 0.994 & 0.9921 \\
Calyx & 0.957 & 0.853 & 0.933 & 0.483 & 0.9018 \\
Stem cavity & 0.827 & 0.656 & 0.789 & 0.313 & 0.7319 \\
\midrule
\textbf{All classes} & \textbf{0.923} & \textbf{0.836} & \textbf{0.906} & \textbf{0.597} & \textbf{0.8775} \\
\bottomrule
\end{tabularx}
\end{table}

\textbf{Discussion.} The combination underperforms both of its constituent configurations on every aggregate metric: overall F1 falls to 0.8775, below both the baseline (0.9047) and the CA-swap-alone model (0.9038), and is the lowest F1 recorded across the entire study. The calyx class does improve slightly relative to CA-swap-alone (F1 0.9018 vs.\ 0.8969), consistent with Run E's isolated calyx benefit, but this gain is purchased at a severe cost to the stem cavity class: recall collapses from 0.872 (CA-swap-alone) to 0.656, pushing stem-cavity F1 to 0.7319 --- below even the untuned baseline (0.7918), and far below the CA-swap model's best-in-study result (0.8217). In other words, combining the two best-performing isolated interventions reverses the architectural factor's primary benefit rather than compounding it. This is consistent with the recall-suppressing pattern observed across several augmentation variants in Section~\ref{sec:augmentation} (Runs E, F, and G all traded recall for precision relative to the baseline); layering that same tendency onto the CA-swap architecture appears to compound against the smallest, most instance-sparse class rather than complementing its architectural gain.

\textbf{Summary of Section~\ref{sec:final_combined}.} This result provides direct empirical confirmation --- rather than inference from separate ablation runs --- that the factors identified as individually beneficial or neutral in this study do not compose additively, and in this case interact destructively. The CA-swap architecture alone (Section~\ref{sec:ca_swap}), without any augmentation modification, is therefore retained as the final recommended configuration; no further combination of ablation factors is pursued.

\subsection{Qualitative Results}
\label{sec:qualitative}

\begin{figure}[!htbp]
\centering
\begin{minipage}[b]{0.32\textwidth}
    \centering
    \includegraphics[width=\textwidth]{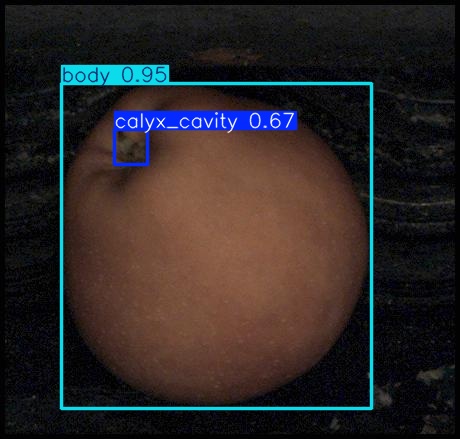}\\[3pt]
    {\normalsize \textbf{(a)}}
\end{minipage}
\hfill
\begin{minipage}[b]{0.32\textwidth}
    \centering
    \includegraphics[width=\textwidth]{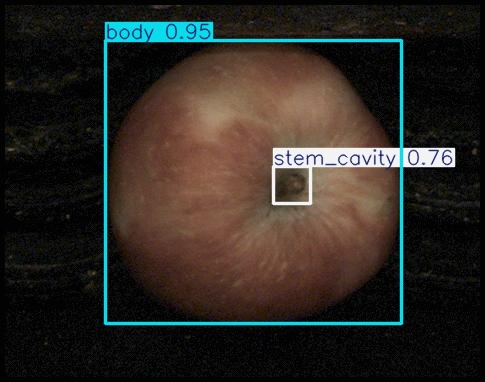}\\[3pt]
    {\normalsize \textbf{(b)}}
\end{minipage}
\hfill
\begin{minipage}[b]{0.32\textwidth}
    \centering
    \includegraphics[width=\textwidth]{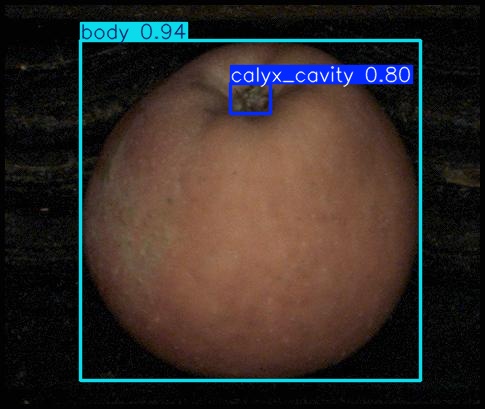}\\[3pt]
    {\normalsize \textbf{(c)}}
\end{minipage}
\caption{Representative detections of the final CA-swap model (at the 0.15 confidence threshold) on held-out test images across varying apple cultivars and skin tones: (\textbf{a}) pale yellow-green apple with detected stem cavity (confidence 0.76); (\textbf{b}) pale pink-blush apple with detected calyx (confidence 0.80); and (\textbf{c}) deep red apple with detected calyx (confidence 0.67).}
\label{fig:qualitative_apples}
\end{figure}

To illustrate how the final recommended model (the CA-swap architecture, Section~\ref{sec:ca_swap}, at the 0.15 confidence threshold selected in Section~\ref{sec:threshold}) performs under the dataset's natural appearance variability, Figure~\ref{fig:qualitative_apples} shows three representative detections on held-out test images spanning the color range described in Section~\ref{sec:dataset}: (a) a pale yellow-green apple with a correctly detected stem cavity (confidence 0.76); (b) a pale pink-blush apple with a correctly detected calyx (confidence 0.80); and (c) a deep red apple with a correctly detected calyx at comparatively lower confidence (0.67), illustrating how darker, more saturated skin tones reduce the model's detection confidence for this class without yet causing a missed detection. All three examples show a correctly localized fruit-body box alongside the landmark detection, consistent with the near-perfect fruit-body performance reported throughout this study (Table~\ref{tab:ca_swap}).

\section{Discussion}
\label{sec:discussion}

Across the five factors examined, a clear pattern emerges: training- and data-level interventions (input resolution, augmentation policy, optimizer/LR schedule) consistently failed to improve on the untuned baseline, while the one architectural intervention tested --- the CoordAtt swap --- was the only factor to match or exceed it, and the only factor to meaningfully improve the stem cavity class. This asymmetry is informative beyond the specific numbers. In a fixed-camera, controlled-illumination acquisition setup, the geometric and photometric variability that augmentation is designed to simulate (viewpoint, lighting, occlusion) is largely already absent from the data distribution the model will see at inference time; synthetic perturbations beyond the Ultralytics default therefore add noise relative to the deployment distribution rather than useful invariance. The optimizer result points in the same direction: with a moderately sized, relatively clean dataset, Ultralytics' automatic optimizer selection is already well matched to the problem, leaving little room for a manually tuned alternative to improve on it. Architecture, by contrast, changes what the model is capable of representing regardless of the data distribution, which is consistent with coordinate attention's specific benefit to the stem cavity class --- the smallest, most texturally variable, and consistently weakest-performing landmark across every other factor tested.

This finding is consistent with, and extends, the two most closely related studies in Section~\ref{sec:relwork_landmark}. Yang et al.'s CA-YOLOv5 \cite{Yang2024} showed that coordinate attention improves whole-apple detection in uncontrolled natural environments; the present results show the same mechanism also improves detection of a specific sub-fruit landmark under controlled imaging conditions, suggesting the benefit of position-sensitive channel attention generalizes across both environmental variability and landmark granularity. Baek et al.'s AS-YOLO \cite{Baek2025} similarly found that an attention mechanism, rather than training-level tuning, was necessary to improve stem-region segmentation --- but reported only a single already-optimized architecture. The controlled, one-factor-at-a-time design used here isolates that same conclusion from first principles: for this class of fine-grained landmark detection problem, architectural modification appears to be a more productive lever than augmentation or optimizer tuning, at least within the ranges tested.

From an industrial deployment perspective, the mechanistic failure of additivity in Section~\ref{sec:final_combined} offers critical guidance for engineering high-speed grading lines. Morphological depressions such as the stem cavity rely on delicate, low-contrast specular shading gradients around the basin edge. While global fruit detection remains robust across extreme color perturbations, aggressive synthetic jittering (Run E) disrupts the spatial-channel co-dependencies that Coordinate Attention explicitly encodes via 1D direction-aware pooling, destroying the subtle shadow contours required to localize the cavity. Furthermore, in mechanical orientation systems utilizing bi-axial roller bobbins or pneumatic flippers, undetected landmarks represent an irreversible sorting failure, forcing fruit recirculation and degrading conveyor throughput. Operating the lighter CA-swap architecture at a calibrated 0.15 threshold maximizes landmark recall while eliminating the quadratic memory overhead of native C2PSA attention blocks, ensuring deterministic, low-latency execution on embedded edge hardware (e.g., NVIDIA Jetson platforms) without sacrificing orientation reliability.

\begin{figure}[!htbp]
\centering
\includegraphics[width=0.75\textwidth]{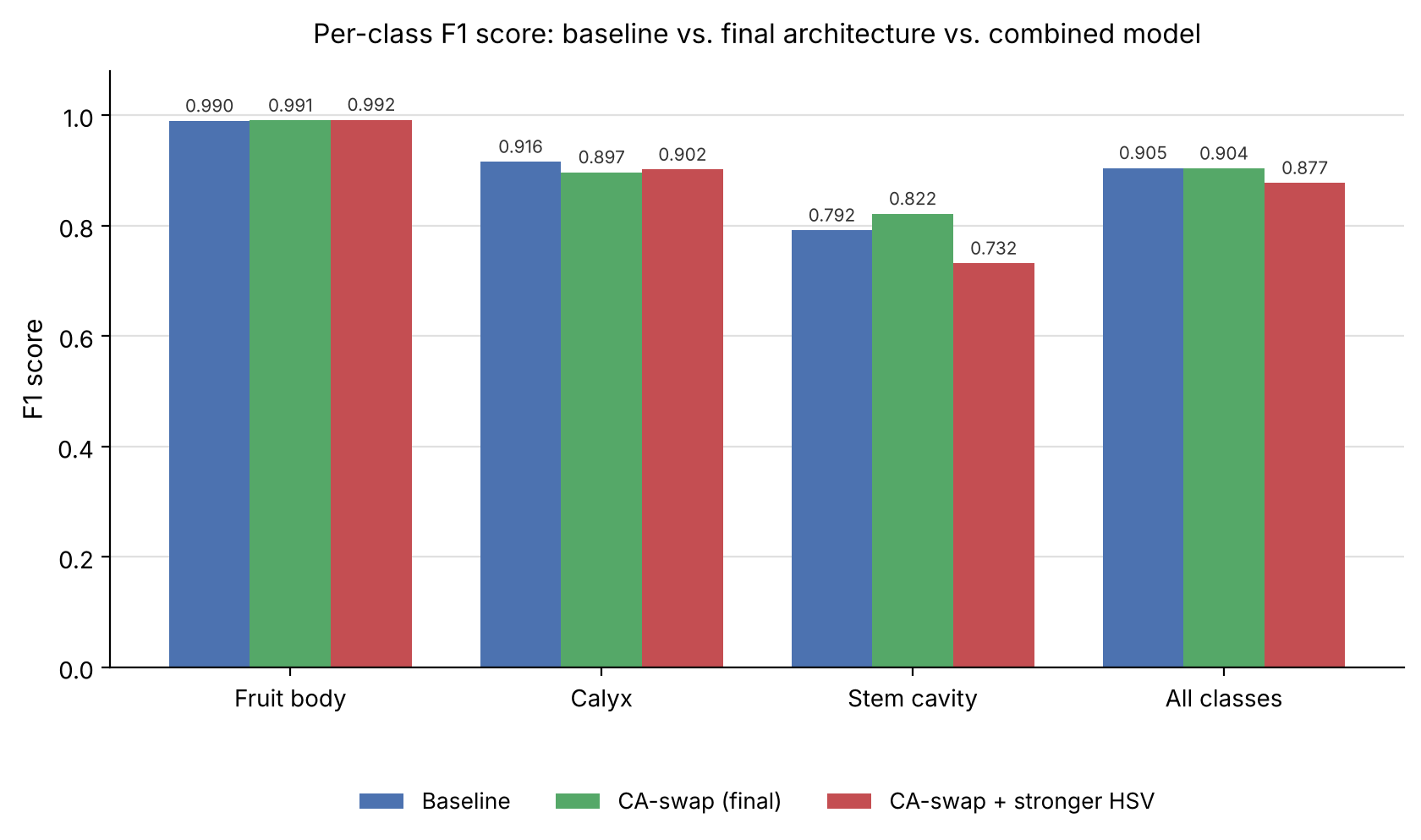}
\caption{Per-class F1 score comparison across the baseline, the final CA-swap architecture (Section~\ref{sec:ca_swap}), and the combined CA-swap-plus-stronger-HSV model (Section~\ref{sec:final_combined}). The architecture's stem-cavity gain and the combination's stem-cavity loss are directly visible as the one class whose bar moves in opposite directions between the second and third configurations, while fruit body remains essentially unchanged across all three.}
\label{fig:ablation_summary}
\end{figure}

Several limitations constrain how far these findings should be generalized. All data were collected under a single controlled, fixed-camera, fixed-illumination setup; whether the same factor rankings hold under variable camera distance, lighting, or an orchard/conveyor environment is untested. The dataset, while fully self-annotated and class-balanced within the constraints of natural fruit morphology, remains moderate in scale (1506 images), and the stem cavity class in particular had the fewest training instances of the three, which may compound its sensitivity to the factors tested. Only one backbone family (YOLO11s) and one attention mechanism (coordinate attention, tested as a single same-slot substitution) were evaluated; other lightweight attention or backbone modifications were not explored and may interact differently with this dataset. Inference-speed measurements also showed session-to-session variance (Section~\ref{sec:resolution}) large enough that the latency figures reported here should be read as indicative rather than tightly reproducible; a dedicated, single-session benchmarking pass across the baseline and final configuration is identified as necessary follow-up work before any latency claim is used for deployment sizing. Finally, the augmentation and optimizer ablations, while systematic, tested a representative rather than exhaustive set of variants within each factor, and only one cross-factor combination (CA-swap with stronger HSV, Section~\ref{sec:final_combined}) was tested directly; it remains possible that a different combination, or a configuration outside the tested range entirely (for example, a more targeted, class-weighted training strategy for the calyx and stem cavity classes, as noted in Section~\ref{sec:augmentation}), could outperform the baseline where the tested variants and this one combination did not.

\section{Conclusions and Future Works}
\label{sec:conclusions}

This study presented a custom, fully self-collected and self-annotated dataset of 1506 apple images annotated for three orientation-relevant landmarks --- fruit body, stem cavity, and calyx --- and used it to conduct a systematic, one-factor-at-a-time ablation study on a YOLO11s baseline, addressing a gap in the apple-detection literature where prior work typically reports a single, already-optimized configuration rather than isolating the contribution of individual design choices. Across five factors --- input resolution, augmentation policy, optimizer/LR schedule, lightweight architectural modules, and inference decision threshold --- the results show that, for this fine-grained, controlled-environment landmark-detection task, architectural modification was the only lever that improved on the untuned baseline, while increased resolution, every tested augmentation variant, and manual optimizer tuning each underperformed it on the primary metrics.

The recommended final configuration combines the default (baseline) augmentation policy, the default automatic optimizer/LR schedule, a coordinate-attention substitution for YOLO11s's native C2PSA block, and a 0.15 confidence threshold at inference. This configuration matches the baseline's overall F1 (0.9038 vs.\ 0.9047) while achieving the study's highest mAP@0.5 (0.935), its best stem-cavity F1 (0.8217), and a roughly 10\% reduction in parameter count and GFLOPs --- a combination of comparable accuracy and reduced computational cost directly relevant to real-time, potentially embedded deployment in industrial apple sorting and orientation systems. Future work should extend this analysis beyond the single controlled imaging environment studied here, benchmark inference latency in a single consistent session on representative embedded hardware, and explore class-targeted training strategies and alternative attention or backbone modules for the stem cavity and calyx classes specifically.

In the future,we would like to implement our deep learning model on mobile platforms. We believe that our CAppleNet will work on mobile platforms at real-time frame rates. Additionally, there is potential for upgrading and optimizing our model by incorporating state of the art deep learning methods and techniques~\cite{Gok2023SIU, Azadvatan2024arXiv, Jabbarli2024arXiv,Akdogan2025ARTEMIS,Turkun2026MDPID,Akdogan2026MDPIM,Azadvatan2026MDPIAS,Cangirili2026arXiv}.

To improve accuracy of CAppleNet, we are also interested in implementing realistic Bidirectional Reflectance Distribution Function (BRDF)~\cite{Ozturk2006EGUK, Kurt2007MScThesis, Ozturk2008CG, Kurt2008SIGGRAPHCG, Kurt2009SIGGRAPHCG, Kurt2010SIGGRAPHCG, Ozturk2010GraphiCon, Szecsi2010SCCG, Ozturk2010CGF, Bigili2011CGF, Bilgili2012SCCG, Ergun2012SCCG, Toral2014SIU, Tongbuasirilai2017ICCVW, Kurt2019DEU, Tongbuasirilai2020TVC, Akleman2024arXiv,Kurt2025arXiv}, Bidirectional Scattering Distribution Function (BSDF)~\cite{WKB12, Ward2014MAM, Kurt2014WLRS, Kurt2016SIGGRAPH, Kurt2017MAM, Kurt2018DEU}, Bidirectional Surface Scattering Reflectance Distribution Function (BSSRDF)~\cite{Kurt2013TPCG, Kurt2013EGSR, Kurt2014PhDThesis, Onel2019PL, Kurt2020MAM, Kurt2021TVC, Yildirim2026MDPIAS} and multi-layered material models~\cite{WKB12, Kurt2016SIGGRAPH, Mir2022DEU} into our framework.

\authorcontributions{Conceptualization, A.G. and M.K.; methodology, A.G. and M.K.; software, A.G.; validation, A.G., R.S.H. and G.H.; formal analysis, A.G. and M.K.; investigation, A.G., R.S.H. and G.H.; resources, R.S.H. and G.H.; data curation, A.G.; writing---original draft preparation, A.G.; writing---review and editing, A.G. and M.K.; visualization, A.G.; supervision, M.K.; project administration, M.K., R.S.H. and G.H. All authors have read and agreed to the published version of the manuscript.}

\funding{This research was funded by the Scientific and Technical Research Council of Turkey grant number 7250234.}

\dataavailability{The custom dataset and implementation source codes used in this study are proprietary to Tarend Agricultural and Industrial Machinery.}

\acknowledgments{The authors express gratitude to Tarend Agricultural and Industrial Machinery for providing the industrial facility, prototype machinery, and testing hardware used during this project.}

\conflictsofinterest{Authors Robert Selim Hill, Gökhan Hasanoğlu and Alican Güneş are employed by Tarend. The remaining author declares no conflict of interest.}

\begin{adjustwidth}{-\extralength}{0cm}

\reftitle{References}

\externalbibliography{yes}
\nocite{*}
\bibliography{MDPI26_CAppleNet_References}

\PublishersNote{}
\end{adjustwidth}
\end{document}